\documentclass[11pt,a4paper]{article}
\usepackage{times}
\usepackage{latexsym}
\usepackage{graphicx}
\usepackage{multirow}
\usepackage{amsmath}

\usepackage{url}

\title{EmotionX-DLC: Self-Attentive BiLSTM for Detecting Sequential Emotions in Dialogues}

\author{Linkai Luo \ \  {\normalfont and} \ \ Haiqin Yang  \ \ {\normalfont and} \ \  Francis Y. L. Chin \ \ \\
	{\tt llk1896@gmail.com, hqyang@ieee.org, francischin@hsmc.edu.hk} \\
	Deep Learning Research and Application Centre \\
	Hang Seng Management College \\
	\\
}

\date{}

\begin{document}
	\maketitle
	\begin{abstract}
		In this paper, we propose a self-attentive bidirectional long short-term memory (SA-BiLSTM) network to predict multiple emotions for the EmotionX challenge. The BiLSTM exhibits the power of modeling the word dependencies, and extracting the most relevant features for emotion classification.  Building on top of BiLSTM, the self-attentive network can model the contextual dependencies between utterances which are helpful for classifying the ambiguous emotions.  We achieve 59.6 and 55.0 unweighted accuracy scores in the \textit{Friends} and the \textit{EmotionPush} test sets, respectively.
	\end{abstract}
	
	\section{Introduction}
	
	Emotion detection plays a crucial role in developing a smart dialogue system such as a chit-chat conversational bot~\cite{DBLP:journals/corr/abs-1802-08379}.  As a typical sub-problem of sentence classification, emotion classification requires not only to  understand sentence of a single utterance, but also capture the contextual information from the whole conversations.
	
	The problems of sentence-level classification have been investigated heavily by means of deep neural networks, such as convolutional neural networks (CNN)~\cite{DBLP:conf/emnlp/Kim14}, long short-term memory (LSTM)~\cite{DBLP:conf/ijcai/LiuQH16}, and attention-based CNN~\cite{DBLP:journals/corr/abs-1804-00831}.  Additional soft attention layers~\cite{DBLP:journals/corr/BahdanauCB14} are usually built on top of those networks, such that more attention will be paid to the most relevant words that lead to a better understanding of the sentence. LSTMs~\cite{DBLP:journals/neco/HochreiterS97} are also useful to model contextual dependencies. For example, a contextual LSTM model is proposed to select the next sentence based on the former context~\cite{DBLP:journals/corr/GhoshVSRDH16}, and a bidirectional LSTM (BiLSTM) is adopted to detect multiple emotions~\cite{DBLP:journals/corr/abs-1802-08379}.
	
	In this work, we utilize the self-attentive BiLSTM (SA-BiLSTM) model to predict multiple types of emotions for the given utterances in the dialogues.  Our model imitates human's two-step procedures for classifying an utterance within the context, i.e., sentence understanding and contextual utterances dependence extraction.  More specifically, we propose the bidirectional long short-term memory (BiLSTM) with the max-pooling architecture to embed the sentence into a fixed-size vector, as the BiLSTM network is capable of modeling the word dependencies in the sentence while the max-pooling helps to reduce the model size and obtains the most related features for emotion classification.  Since data in this challenge is limited and specific words play significant role to classifying the corresponding emotion, we apply the self-attention network~\cite{DBLP:conf/nips/VaswaniSPUJGKP17} to extract the dependence of all the utterances in the dialogue.  Technically, the self-attention model computes the influence of utterance pairs and outputs the sentence embedding of one utterance by a weighted  sum over all the utterances in the dialogue.  The fully connected layers are then applied on the output sentence embedding to classify the corresponding emotion.
	
	\begin{figure}[!hbt]
	\centering
	\includegraphics[scale=0.45]{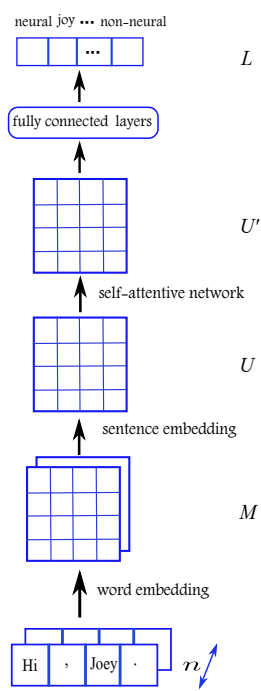}
	\caption{The model architecture illustration.  A bunch of $n$ utterances in one dialogue are processed through word embedding, sentence embedding, self-attentive and fully connected layers.}
	\label{fig:model}
	\end{figure}
	
	\section{Model}
	Figure~\ref{fig:model} presents our designed model architecture.  First, the pre-trained 300-dimensional GloVe vectors~\cite{DBLP:conf/emnlp/PenningtonSM14} are adopted to represent each word (token).  A sentence (utterance) with $m$ tokens is then represented by
	\begin{equation}
	S = (w_1, w_2 \dots, w_m), 
	\end{equation}
	where $w_i$ is $d$-dimensional word embedding for the $i$-th tokens in the sentence.  
	
	Suppose a dialogue consists of $n$ sentences, the input forms an $n \times m \times d$ tensor, $M$, see Figure~\ref{fig:model}.  Via the process of sentence embedding (elaborated in Section~\ref{sec:encoder}), the tensor is converted to a $n\times 2l$ matrix $U$, where $l$ is the number of the hidden units for each unidirectional LSTM.  By applying the self-attentive network, we re-weight the sentence embedding matrix to $U'$ with the same shape as $U$.  Finally, fully connected layers are trained to establish the mapping between input $U'$ and the output emotion labels.
	
	\begin{figure}[!hbt]
		\centering
		\includegraphics[scale=0.35]{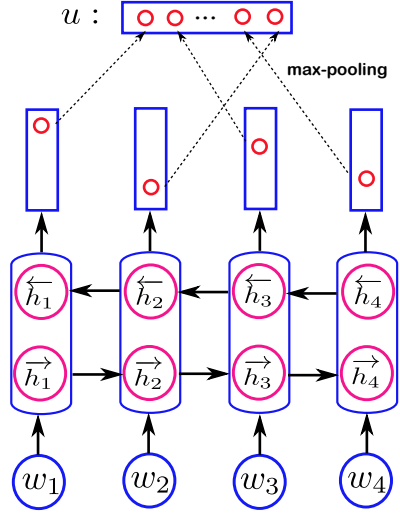}
		\caption{BiLSTM with max-pooling network.}
		\label{fig:bilstm}
	\end{figure}
	
	\subsection{Sentence Embedding}
	\label{sec:encoder}
	In this work, we adopt the BiLSTM to learn the sentence embedding because it is the most popular neural network architecture to encode sentences~\cite{DBLP:conf/emnlp/ConneauKSBB17, DBLP:journals/corr/LinFSYXZB17}.  The forward LSTM and backward LSTM read the sentence $S$ in two opposite directions (see Figure~\ref{fig:bilstm}):
	\begin{align}
	\overrightarrow{h}_t &= \overrightarrow{LSTM}\left( w_t, \overrightarrow{h}_{t-1} \right) \\
	\overleftarrow{h}_t &= \overleftarrow{LSTM}\left( w_t, \overleftarrow{h}_{t+1} \right)
	\end{align}
	The vectors $\overrightarrow{h}_t$ and $\overleftarrow{h}_t$ are concatenated to a hidden state $h_t$.  Max-pooling~\cite{DBLP:conf/icml/CollobertW08} is then conducted along all the words of a sentence to output the final sentence representation, $u$.
	
	\subsection{The Self-attentive Network}
	
	\begin{figure}[!hbt]
		\centering
		\includegraphics[scale=0.38]{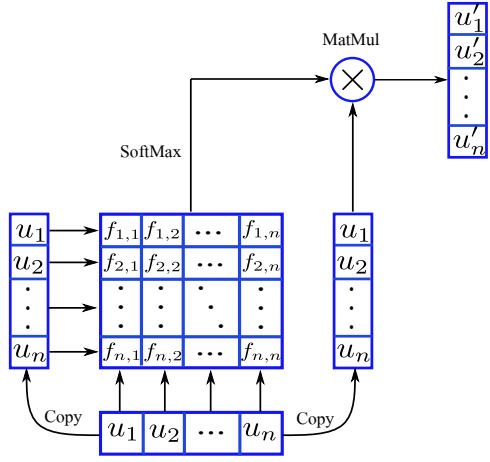}
		\caption{The self-attentive network. $f_{ij}$ denotes $f(u_i, u_j)$ in Eq.~\ref{eq:attn}.}
		\label{fig:self-attention}
	\end{figure}
	%The Transformer is built solely on attention mechanism, or dot-product attention, more specifically. The selfattentive network we adopt in this work is shown in Figure~\ref{fig:self-attention}. Given a query matrix $Q$, key matrix $K$ and value matrix $V$ (in our case the $Q$, $K$, $V$ are the identical matrix $U$), the attention weight is calculated by:
	
	The self-attention model, called Transformer~\cite{DBLP:conf/nips/VaswaniSPUJGKP17}, is an effective non-recurrent architecture for machine translation.  We adopt it to capture the utterances dependence.  Figure~\ref{fig:self-attention} shows the model to build the dot-product attention of utterances, where the attention matrix is calculated by:
	\begin{equation}\label{eq:attn}
	f(u_i, u_j) = \left\{ \begin{array}{ll}
	\frac{u_iu_j^T}{\sqrt{d_k}}, & \mbox{if $i,j \leq n$};\\
	-\infty & \mbox{otherwise}.\end{array} \right. 
	\end{equation}
	where $u_i$ is the $i$-th sentence embedding in the dialogue, and $d_k$ (i.e. $2l$) is the model dimension. An attention mask is applied to waive the inner attention between sentence embeddings and paddings.  
	% \mathrm{softmax} 
	
	The $i$-th sentence embedding is finally weighted by summing over all the sentence embeddings to enhance the effect:
	\begin{equation}
	u_i' = \sum_j \mathrm{softmax}\left( F_i \right)\, u_j,
	\end{equation} 
	where $F_i$ is $n$-dimensional vector whose $j$-th element is $f(u_i, u_j)$.
	\if 0
	We also apply attention mask to waiver the inner attention between sentence embeddings and paddings (we add paddings to the dialogue for null sentences):
	\begin{equation}
	f_{ij} = \left\{ \begin{array}{ll}
	f_{ij} & \mbox{if $i,j \leq n$};\\
	-\infty & \mbox{otherwise}.\end{array} \right.
	\end{equation}
	\fi 
	
	\subsection{Output and Loss}
	Finally, we apply fully connected layers to produce the corresponding emotions.  In the training, the weighted cross-entropy is adopted as the loss function.  Since the challenge only focuses on classifying four types of emotions, rather than all eight types, we set the weights to zero for the unconsidered emotions.
	
	\section{Experiments and Results}
	\subsection{Data Preprocessing}
	The EmotionX dataset consists of the \textit{Friends} TV scripts and the \textit{EmotionPush} chat logs on Faceook Messenger in eight types, i.e., Neutral, Joy, Sadness, Anger, Fear, Surprise, Disgust and Non-neutral.  For the train set, there are 720 dialogues for \textit{Friends} and \textit{EmotionPush}, respectively, which yields a total of 1,440 dialogues, 21,294 sentences, and 9,885 unique words.  In the challenge, we test the following candidate labels, Neutral, Joy, Sadness, and Anger.  We also conduct the following steps to clean the data:
	\begin{itemize}
		\item Unicode symbols, except emojis (the direct expressions of human emotions), are removed.  Person names, locations, numbers and websites are replaced with special tokens.
		\item The Emoji symbols are converted to the corresponding meanings.
		\item Duplicated punctuation and symbols. Tokens with duplicated punctuation or alphabets, such as ``oooooh", often imply non-neural emotions.  We reconstruct the tokens to be \texttt{oh <duplicate>} to avoid informal words. The same rule also applies to similar tokens.  For example, ``oh!!!!!!" is replaced by \texttt{oh ! <duplicate>}.
		\item Word tokenization. We use NLTK's TwitterTokenizer~\cite{DBLP:conf/acl/Bird06} to split the sentences into tokens. All tokens are set lowercase.
	\end{itemize}
	
	\subsection{Experimental Setup}
	We conduct two experiments with different model variants: BiLSTM and SA-BiLSTM, to validate whether our proposed model can learn the contextual information. The network settings for each model are summarized as follows:
	\begin{itemize}
		\item BiLSTM: BiLSTM + max-pooling + fully connected layers.
		\item SA-BiLSTM: BiLSTM + max-pooling + self-attentive network + fully connected layers.
	\end{itemize}
	The word embedding is 300-dimensional from the the Glove.  Pack padded sequence and pad packed sequence are implemented to deal with varying sequence lengths.  For SA-BiLSTM, we limit the utterance number to 25 for each dialogue.  Due to the limit of training data, LSTM is set to one layer with only 256 hidden units.  The fully connected layers consist of two middle layers with the same size of 128.  
	
	The mini-batch size for training BiLSTM is set to 16.  Unlike BiLSTM, we feed one dialogue to SA-BiLSTM for every training step.  Adam~\cite{DBLP:journals/corr/KingmaB14} is the adopted optimizer with initial learning rate 0.0002 and decay factor 0.99 for every epoch.  Dropout probability is set to 0.3 for BiLSTM and self-attention layers.  We train BiLSTM for 10 epochs and SA-BiLSTM for 20 epochs to gain the best accuracy in the validation sets.
	
	\begin{table*}[!hbt]
		\centering
		\begin{tabular}{l|l|c|c|*{4}{c}}
			\hline \hline
			Model & Dataset & WA & UWA &Neutral &Joy &Sadness &Anger\\ \hline
			\multirow{2}{*}{BiLSTM} & \textit{Friends} &\bf 79.4 &60.4 &92.5 &78.9 &29.0 &41.2\\ 
			& \textit{EmotionPush} &\bf 83.9 &61.8 &92.6 &73.1 &41.0 &40.4 \\ \hline 
			\multirow{2}{*}{SA-BiLSTM} & \textit{Friends} &78.8 &\bf62.8 &90.6 &73.2 &40.3 &47.1 \\ 
			& \textit{EmotionPush} &83.4 &\bf63.5 &91.9 &69.6 &47.0 &45.7\\ \hline \hline
		\end{tabular}
		\caption{Experimental results of \textit{Friends} and \textit{EmotionPush} in the validation sets.}
		\label{tb:result}
	\end{table*}
	
	\begin{table*}[!hbt]
		\centering
		\begin{tabular}{l|l|c|*{4}{c}}
			\hline \hline
			Model & Dataset & UWA &Neutral &Joy &Sadness &Anger \\ \hline
			\multirow{3}{*}{SA-BiLSTM} & \textit{Friends}  &59.6 &90.1 &68.8 &30.6 &49.1 \\ 
			& \textit{EmotionPush} &55.0 &94.2 &70.5 &31.0 &24.3 \\ \cline{2-7}
			& \textit{Average} &57.3 &92.1 &69.6 &30.8 &36.7 \\ \hline \hline
		\end{tabular}
		\caption{Experimental results of \textit{Friends} and \textit{EmotionPush} in the test sets.}
		\label{tb:result2}
	\end{table*}
	
	\subsection{BiLSTM Versus SA-BiLSTM}
	
	Table~\ref{tb:result} reports the model performance in the validation sets, which consist of 80 dialogues for \textit{Friends} and \textit{EmotionPush}, respectively.  We evaluate two criteria, the weighted accuracy (WA) and the unweighted accuracy (UWA)~\cite{DBLP:journals/corr/abs-1802-08379}. The predicted accuracy for each class is also given in the table.
	
	%Note that our model focuses only on the prediction among four candidate emotions, i.e., neutral, joy, sadness and anger. In general, all the models perform better for \textit{EmotionPush} than \textit{Friends}, because the former dialogues are simpler in terms of short utterance length, less complicated language and smaller vocabulary. Compared with the existing models, it is not surprised to see that both WA and UWA of our models have dramatically improved over CNN and CNN-BiLSTM that were built on seven classes of emotions. The maximum WA were achieved by BiLSTM, and UWA by SA-BiLSTM. Nevertheless, the comparison is not fair at this moment; firstly, the models have different focuses and second, our models were only evaluated on the validation set.
	
	Interestingly, the simpler model BiLSTM achieves higher WA, with up to 0.6\% and 0.5\% improvement in \textit{Friends} and \textit{EmotionPush}, respectively.  On the other hand, SA-BiLSTM overperforms BiLSTM in terms of UWA, with up to 1.4\% and 1.7\% improvement.  Note that BiLSTM tends to predict the emotions Neutral and Joy far more accurate than the other two emotions because most utterances are labeled as these two emotions, i.e., 45.03\% as neural and 11.79\% as joy in \textit{Friends} while 66.85\% as neural and 14.25\% as joy in \textit{EmotionPush}.  Overall, SA-BiLSTM provides a more balanced prediction for each type of emotion than BiLSTM.  Especially in predicting the emotions of Sadness and Anger, SA-BiLSTM gains better predictive accuracy, up to 11.3\% \& 6.0\% on the Sadness emotion and 5.9\% \& 5.3\% on the Anger emotion improvements in \textit{Friends} and \textit{EmotionPush}, respectively.
	
	\subsection{Results of Test Set}
	We submit the results produced by SA-BiLSTM and obtain the evaluation scores provided by the challenge organizer.  Table~\ref{tb:result2} lists the experimental results evaluated in the test set, which consists of 400 dialogues, 200 dialogues for \textit{Friends} and \textit{EmotionPush}, respectively.
	
	The results indicate that our model shows a strong bias towards predicting the Neutral emotion and the Joy emotion in both datasets compared to the Sadness and the Anger emotions.  Especially, our model achieves an extremely poor prediction on the Anger emotion in \textit{EmotionPush}.  Moreover, the UWA in \textit{EmotionPush} is smaller than that in \textit{Friends}, which is different from our prediction results in the validation set.  We conjecture that the distribution of the validation set and the test set may be slightly different.  To obtain a robust solution, we may train multiple models using different random seeds and ensemble the model averaged on the checkpoints.  
	
	We notice that the \emph{Speaker} information is also important for emotion classification. Table~\ref{tb3} shows two consecutive utterances made by the same speaker from \textit{EmotionPush}, where the first utterance seems literally less emotional than the second one. Nevertheless, the two utterances should carry the same emotion, i.e., Anger,  because they are made by the same speaker consecutively. On the contrary, our model gives a false prediction (i.e., Neutral) for the second utterance because it probably treats the two utterance separately. We believe that our model shall gain some improvements by adding speaker information into it.
	
	\begin{table}[!hbt]
	\centering
	\begin{tabular}{lp{4.8cm}}

		\bf Speaker ID &\bf Utterance \\ \hline
		1051336806 & \it but /you/ bug /me/ \\
		1051336806 & \it and you hundred percent told peopel stfu.  \\ \hline
	\end{tabular}
	\caption{Consecutive utterances made by the same speaker shall carry the same emotion.}
	\label{tb3}
	\end{table}
	
	\section{Conclusion}
	
	In this work, we propose SA-BiLSTM to predict multiple emotions for given utterances in the dialogues.  The proposed network is a self-attentive network built on top of BiLSTM.  Our results evaluated on the validation set show that BiLSTM has better WA performance, while SA-BiLSTM is advantageous to BiLSTM in terms of UWA.  According to the test results, SA-BiLSTM yields higher UWA scores for detecting the Neural emotion and the Joy emotions than the Sadness and the Anger ones.  The bias may be caused by uneven training data distributions.  We hope to improve our model by either incorporating more related data or retrieving more linguistic information.
	
	\section*{Acknowledgment}
	This work was partially supported by the Research Grants Council of the Hong Kong Special Administrative Region, China (Project No. UGC/IDS14/16).
	
%	\bibliographystyle{unsrtnat}
    \bibliographystyle{abbrv}
	\bibliography{emotionX}

\begin{thebibliography}{10}

\bibitem{DBLP:journals/corr/BahdanauCB14}
D.~Bahdanau, K.~Cho, and Y.~Bengio.
\newblock Neural machine translation by jointly learning to align and
  translate.
\newblock {\em CoRR}, abs/1409.0473, 2014.

\bibitem{DBLP:conf/acl/Bird06}
S.~Bird.
\newblock {NLTK:} the natural language toolkit.
\newblock In {\em {ACL} 2006, 21st International Conference on Computational
  Linguistics and 44th Annual Meeting of the Association for Computational
  Linguistics, Proceedings of the Conference, Sydney, Australia, 17-21 July
  2006}, 2006.

\bibitem{DBLP:journals/corr/abs-1802-08379}
S.~Chen, C.~Hsu, C.~Kuo, T.~Huang, and L.~Ku.
\newblock Emotionlines: An emotion corpus of multi-party conversations.
\newblock {\em CoRR}, abs/1802.08379, 2018.

\bibitem{DBLP:conf/icml/CollobertW08}
R.~Collobert and J.~Weston.
\newblock A unified architecture for natural language processing: deep neural
  networks with multitask learning.
\newblock In {\em Machine Learning, Proceedings of the Twenty-Fifth
  International Conference {(ICML} 2008), Helsinki, Finland, June 5-9, 2008},
  pages 160--167, 2008.

\bibitem{DBLP:conf/emnlp/ConneauKSBB17}
A.~Conneau, D.~Kiela, H.~Schwenk, L.~Barrault, and A.~Bordes.
\newblock Supervised learning of universal sentence representations from
  natural language inference data.
\newblock In {\em Proceedings of the 2017 Conference on Empirical Methods in
  Natural Language Processing, {EMNLP} 2017, Copenhagen, Denmark, September
  9-11, 2017}, pages 670--680, 2017.

\bibitem{DBLP:journals/corr/GhoshVSRDH16}
S.~Ghosh, O.~Vinyals, B.~Strope, S.~Roy, T.~Dean, and L.~P. Heck.
\newblock Contextual {LSTM} {(CLSTM)} models for large scale {NLP} tasks.
\newblock {\em CoRR}, abs/1602.06291, 2016.

\bibitem{DBLP:journals/neco/HochreiterS97}
S.~Hochreiter and J.~Schmidhuber.
\newblock Long short-term memory.
\newblock {\em Neural Computation}, 9(8):1735--1780, 1997.

\bibitem{DBLP:conf/emnlp/Kim14}
Y.~Kim.
\newblock Convolutional neural networks for sentence classification.
\newblock In {\em Proceedings of the 2014 Conference on Empirical Methods in
  Natural Language Processing, {EMNLP} 2014, October 25-29, 2014, Doha, Qatar,
  {A} meeting of SIGDAT, a Special Interest Group of the {ACL}}, pages
  1746--1751, 2014.

\bibitem{DBLP:journals/corr/abs-1804-00831}
Y.~Kim, H.~Lee, and K.~Jung.
\newblock Attnconvnet at semeval-2018 task 1: Attention-based convolutional
  neural networks for multi-label emotion classification.
\newblock {\em CoRR}, abs/1804.00831, 2018.

\bibitem{DBLP:journals/corr/KingmaB14}
D.~P. Kingma and J.~Ba.
\newblock Adam: {A} method for stochastic optimization.
\newblock {\em CoRR}, abs/1412.6980, 2014.

\bibitem{DBLP:journals/corr/LinFSYXZB17}
Z.~Lin, M.~Feng, C.~N. dos Santos, M.~Yu, B.~Xiang, B.~Zhou, and Y.~Bengio.
\newblock A structured self-attentive sentence embedding.
\newblock {\em CoRR}, abs/1703.03130, 2017.

\bibitem{DBLP:conf/ijcai/LiuQH16}
P.~Liu, X.~Qiu, and X.~Huang.
\newblock Recurrent neural network for text classification with multi-task
  learning.
\newblock In {\em Proceedings of the Twenty-Fifth International Joint
  Conference on Artificial Intelligence, {IJCAI} 2016, New York, NY, USA, 9-15
  July 2016}, pages 2873--2879, 2016.

\bibitem{DBLP:conf/emnlp/PenningtonSM14}
J.~Pennington, R.~Socher, and C.~D. Manning.
\newblock Glove: Global vectors for word representation.
\newblock In {\em Proceedings of the 2014 Conference on Empirical Methods in
  Natural Language Processing, {EMNLP} 2014, October 25-29, 2014, Doha, Qatar,
  {A} meeting of SIGDAT, a Special Interest Group of the {ACL}}, pages
  1532--1543, 2014.

\bibitem{DBLP:conf/nips/VaswaniSPUJGKP17}
A.~Vaswani, N.~Shazeer, N.~Parmar, J.~Uszkoreit, L.~Jones, A.~N. Gomez,
  L.~Kaiser, and I.~Polosukhin.
\newblock Attention is all you need.
\newblock In {\em Advances in Neural Information Processing Systems 30: Annual
  Conference on Neural Information Processing Systems 2017, 4-9 December 2017,
  Long Beach, CA, {USA}}, pages 6000--6010, 2017.

\end{thebibliography}
	
\end{document}